\documentclass[11pt,a4paper]{article}
\usepackage[hyperref]{acl2019}
\usepackage{times}
\usepackage{url}
\usepackage{latexsym}
\usepackage{url}

\usepackage{amsmath}
\usepackage{amsfonts}
\usepackage{mathtools}
\usepackage{url}
\usepackage{booktabs}
\usepackage{longtable}
\usepackage{graphicx}

\newcommand{\bleupercent}{$\textsc{Bleu}^{[\%]}$}
\newcommand{\bleu}{\textsc{Bleu}}

\newcommand{\ppl}{\textsc{Ppl}}
\newcommand{\ssig}[1]{\phantom{*}$\text{#1}^{*}$}

\newcommand\blfootnote[1]{%
  \begingroup
  \renewcommand\thefootnote{}\footnote{#1}%
  \addtocounter{footnote}{-1}%
  \endgroup
}

\aclfinalcopy 
\def\aclpaperid{10} 

\title{Generalizing Back-Translation in Neural Machine Translation}

\author{Miguel Gra\c{c}a$^{1\dagger}$ \hspace{7pt} Yunsu Kim$^{1}$ \hspace{7pt} Julian Schamper$^{1\dagger}$ \hspace{7pt} Shahram Khadivi$^{2}$ \hspace{7pt} Hermann Ney$^{1}$\\
  $^{1}$Human Language Technology and Pattern Recognition Group \\ 
  RWTH Aachen University, Aachen, Germany\\
  {\tt \{surname\}@i6.informatik.rwth-aachen.de}\\
  $^{2}$eBay, Inc., Aachen, Germany\\
  {\tt \{skhadivi\}@ebay.com}\\}
\fi

\date{}

\begin{document}
\maketitle
\begin{abstract}
Back-translation --- data augmentation by translating target monolingual data --- is a crucial component in modern neural machine translation (NMT). 
In this work, we reformulate back-translation in the scope of cross-entropy optimization of an NMT model, clarifying its underlying mathematical assumptions and approximations beyond its heuristic usage. 
Our formulation covers broader synthetic data generation schemes, including sampling from a target-to-source NMT model.
With this formulation, we point out fundamental problems of the sampling-based approaches and propose to remedy them by (i) disabling label smoothing for the target-to-source model and (ii) sampling from a restricted search space.
Our statements are investigated on the WMT 2018 German $\leftrightarrow$ English news translation task.
\end{abstract}

\blfootnote{$\dagger$ Now at DeepL GmbH.}

\section{Introduction}
\label{s:introduction}

Neural machine translation (NMT) \cite{bahdanau2014neural,vaswani2017attention} systems make use of back-translation \cite{sennrich2016improving} to leverage monolingual data during the training.
Here an inverse, target-to-source, translation model generates synthetic source sentences, by translating a target monolingual corpus, which are then jointly used as bilingual data. 

Sampling-based synthetic data generation schemes were recently shown to outperform beam search \cite{edunov2018understanding,imamura2018enhancement}.
However, the generated corpora are reported to stray away from the distribution of natural data \cite{edunov2018understanding}.
In this work, we focus on investigating why sampling creates better training data by re-writing the loss criterion of an NMT model to include a model-based data generator. 
By doing so, we obtain a deeper understanding of synthetic data generation methods, identifying their desirable properties and clarifying the practical approximations.

In addition, current state-of-the-art NMT models suffer from probability smearing issues \cite{ott2018analyzing} and are trained using label smoothing \cite{pereyra2017regularizing}.
These result in low-quality sampled sentences, which influence the synthetic corpora.
We investigate considering only high-quality hypotheses by restricting the search space of the model via (i) ignoring words under a probability threshold during sampling and (ii) $N$-best list sampling.

We validate our claims in experiments on a controlled scenario derived from the WMT 2018 German $\leftrightarrow$ English translation task, which allows us to directly compare the properties of synthetic and natural corpora.
Further, we present the proposed sampling techniques on the original WMT German $\leftrightarrow$ English task.
The experiments show that our restricted sampling techniques work comparable or superior to other generation methods by imitating human-generated data better.
In terms of translation quality, these do not result in consistent improvements over the typical beam search strategy.

\section{Related Work}
\label{s:related-work}

\newcite{sennrich2016improving} introduce the back-translation technique for NMT and show that the quality of the back-translation model, and therefore resulting pseudo-corpus, has a positive effect on the quality of the subsequent source-to-target model. 
These findings are further investigated in \cite{hoang2018iterative,burlot2018using} where the authors confirm work effect.
In our work, we expand upon this concept by arguing that the quality of the resulting model not only depends on the data fitness of the back-translation model but also on how sentences are generated from it.

\newcite{cotterell2018explaining} frame back-translation as a variational process, with the space of source sentences as the latent space. 
Their approach argues that the distribution of the synthetic data generator and the true translation probability should match.
Thus it is invaluable to clarify and investigate the sampling distributions that current state-of-the-art data generation techniques utilize.
A simple property is that a target sentence must be allowed to be aligned to multiple source sentences during the training phase.
Several efforts \cite{hoang2018iterative,edunov2018understanding,imamura2018enhancement} confirm that this is in fact beneficial. 
Here, we unify these findings by re-writing the optimization criterion of NMT models to depend on a data generator, which we define for beam search, sampling and $N$-best list sampling approaches.

\section{How Back-Translation Fits in NMT}
\label{s:backtranslation}

In NMT, one is interested in translating a source sentence $f_1^J = f_1, \dots, f_j, \dots, f_J$ into a target sentence $e_1^I = e_1, \dots, e_i, \dots,  e_I$. 
For this purpose, the translation process is modelled via a neural model $p_{\theta}(e_i | f_1^J, e_1^{i-1})$ with parameters $\theta$.

The optimal optimization criterion of an NMT model requires access to the true joint distribution of source and target sentence pairs $Pr(f_1^J, e_1^I)$.
This is approximated by the empirical distribution $\hat{p}(f_1^J, e_1^I)$ derived from a bilingual data-set $(f_{1,s}^{J_s}, e_{1,s}^{I_s})_{s=1}^S$.
The model parameters are trained to minimize the cross-entropy, normalized over the number of target tokens, over the same.
\begin{align}
     L(\theta)  &= -  \smashoperator[r]{\sum_{(f_1^J, e_1^I)}} Pr(f_1^J, e_1^I) \cdot \frac{1}{I} \log p_{\theta} \big (e_1^I |f_1^J)  \label{eq:nmt-training1} \\
                &= -  \smashoperator[r]{\sum_{(f_1^J, e_1^I)}} \hat{p}(f_1^J, e_1^I) \cdot \frac{1}{I} \log p_{\theta} \big (e_1^I |f_1^J) \label{eq:nmt-training2} \\
                &= - \frac{1}{S} \sum_{s=1}^S \frac{1}{I_s} \log p_{\theta} \big (e_{1,s}^{I_s} |f_{1,s}^{J_s}) \label{eq:nmt-training3}
\end{align}

Target monolingual data can be included by generating a pseudo-parallel source corpus via, e.g. back-translation or sampling-based methods.
In this section, we describe such generators as a component of the optimization criterion of NMT models and discuss approximations made in practice.

\subsection{Derivation of the Generation Criterion}
\label{s:backtranslation:opt-criterion}

Eq. \ref{eq:nmt-training1} is the starting point of our derivation in Eqs. \ref{eq:nmt-train-opt:1}-\ref{eq:nmt-train-opt:3}.
$Pr(f_1^J, e_1^I)$ can be decomposed into the true language probability $Pr(e_1^I)$ and true translation probability $Pr(f_1^J | e_1^I)$.
These two probabilities highlight the assumptions in the scenario of back-translation: we have access to an empirical target distribution $\hat{p}(e_1^I)$ with which $Pr(e_1^I)$ is approximated, derived from the monolingual corpus $(e_{1,s}^{I_s})_{s=1}^S$.
However, one lacks access to $\hat{p}(f_1^J | e_1^I)$. 
Generating synthetic data is essentially the approximation of the true probability of $Pr(f_1^J | e_1^I)$.
It can be described as a sampling distribution\footnote{The properties of a probability distribution hold for $q(f_1^J | e_1^I; p)$.} $q(f_1^J | e_1^I; p)$ parameterized by the target-to-source model $p$.

\begin{small}
\begin{align}
&L(\theta) = - \smashoperator[r]{\sum_{(f_1^J, e_1^I)}} Pr(f_1^J, e_1^I) \cdot \frac{1}{I} \log p_{\theta} \big (e_1^I |f_1^J) \label{eq:nmt-train-opt:1}\\
&= - \smashoperator[r]{\sum_{e_1^I}} Pr(e_1^I) \cdot \frac{1}{I} \smashoperator[r]{\sum_{f_1^J}} Pr(f_1^J | e_1^I)  \cdot \log  p_{\theta} \big (e_1^I |f_1^J)  \\
   &= - \smashoperator[r]{\sum_{e_1^I}} \hat{p}(e_1^I) \cdot \frac{1}{I} \smashoperator[r]{\sum_{f_1^J}} q(f_1^J | e_1^I; p)  \cdot \log  p_{\theta} \big (e_1^I |f_1^J) \label{eq:nmt-train-opt:3} 
\end{align}
\end{small}

This derivation highlights an apparent condition that the generation procedure $q(f_1^J | e_1^I; p)$ should result in a distribution of source sentences similar to the true data distribution $Pr(f_1^J | e_1^I)$. 
\newcite{cotterell2018explaining} show a similar derivation hinting towards an iterative wake-sleep variational scheme \cite{hinton1995wake}, which reaches similar conclusions. 

Following this, we formulate two issues with the back-translation approach: (i) the choice of generation procedure $q$ and (ii) the adequacy of the target-to-source model $p$.
The search method $q$ is responsible not only for controlling the output of source sentences but also to offset the deficiencies of the target-to-source model $p$.

An implementation for $q$ is, for example, beam search where $q$ is a \emph{deterministic} sampling procedure, which returns the highest scoring sentence according to the search criterion:
\begin{equation}
\label{eq:beam-search-sampling}
\begin{split}
&q_{\text{beam}}(f_1^J | e_1^I; p) = \\
&\Bigg\{\begin{array}{ll}
                    1, & f_1^J = \underset{\hat{J}, f_1^{\hat{J}}}{\text{argmax}} \Big \{ \frac{1}{\hat{J}} \log p(f_1^{\hat{J}} | e_1^{I}) \Big \} \\
                    0, & \text{otherwise } \\ 
                    \end{array} 
\end{split}
\end{equation}

Sampling as described by \newcite{edunov2018understanding} would be simply the equality
\begin{equation}
    q_{\text{sample}}(f_1^J | e_1^I; p) = p(f_1^J | e_1^I).\vspace{0.5em}
\end{equation}

\subsection{Approximations}
\label{s:backtranslation:approximations}

Applications of back-translation and its variants largely follows the initial approach presented in \cite{sennrich2016improving}.
Each target authentic sentence is aligned to a single synthetic source sentence. 
This new dataset is then used as if it were bilingual.
This section is dedicated to the clarification of the effect of such a strategy in the optimization criterion, especially with non-deterministic sampling approaches  \cite{edunov2018understanding,imamura2018enhancement}.

Firstly, the sum over all possible source sentences in Eq. \ref{eq:nmt-train-opt:3} is approximated by a restricted search space of $N$ sentences, with $N=1$ being a common choice.
Yet, the cost of \emph{generating} the data and \emph{training} on the same scales linearly with $N$ and it is unattractive to choose higher values.

Secondly, the pseudo-corpora are static across training, i.e. the synthetic sentences do not change across training epochs, which appears to cancel out the benefits of sampling-based methods.
Correcting this behaviour requires an on-the-fly sentence generation, which increases the complexity of the implementation and slows down training considerably.
Back-translation is not affected by this approximation since the target-to-source model always generates the same translation.

The approximations are shown in Eq. \ref{eq:nmt-approx} with a fixed pseudo-parallel corpus where $e_{1,s}^{I_s}$ is aligned to $N$ source sentences $(f_{1,s,n}^{J_{s},n})_{n=1}^N$.
\begin{align}
\label{eq:nmt-approx}
  &L(\theta) \approx - \smashoperator[r]{\sum_{s=1}^S} \frac{1}{N \cdot I_s}  \smashoperator[r]{\sum_{n=1}^N}  \log  p_{\theta} \big (e_{1,s}^{I_s} | f_{1,s,n}^{J_{s},n}) 
\end{align}

We hypothesize that these conditions become less problematic when large amounts of monolingual data are present due to the law of large numbers, which states that repeated occurrences of the same sentence $e_1^I$ will lead to a representative distribution of source sentences $f_1^J$ according to $q(f_1^J | e_1^I; p)$.
In other words, given a high number of representative target samples, Eq. \ref{eq:nmt-approx} matches Eq. \ref{eq:nmt-train-opt:3} with $N=1$.
This shifts the focus of the problem to find an appropriate search method $q$ and generator $p$.

\section{Improving Synthetic Data}
\label{s:synthetic-data}

In this section, we discuss how the known generation methods $q(f_1^J | e_1^I; p)$ fail in approximating $Pr(f_1^J | e_1^I)$ due to modelling issues of model $p$ and consider how the generation approach $q$ can be adapted to compensate $p$.

We base our remaining work on the approximations presented in Section \ref{s:backtranslation:approximations} and consider $N=1$ synthetic sentences.
The reasoning for this is two-fold: (i) it is the most attractive scenario in terms of computational costs and (ii) the approximations lose their influence with large target monolingual corpora.

\subsection{Issues in Translation Modelling}
\label{s:synthetic-data:issues}

With sampling-based approaches, one does not only care about whether high-quality sentences get assigned a high probability, but also that low-quality sentences are assigned a low probability. 

Label smoothing (LS) \cite{pereyra2017regularizing} is a common component of state-of-the-art NMT systems \cite{ott2018analyzing}.
This teaches the model to (partially) fit a uniform word distribution, causing unrestricted sampling to periodically sample from the same.
Even without LS, NMT models tend to smear their probability to low-quality hypotheses \cite{ott2018analyzing}.

\begin{figure}[t]
\begin{center}
\scalebox{1.0}{
\includegraphics{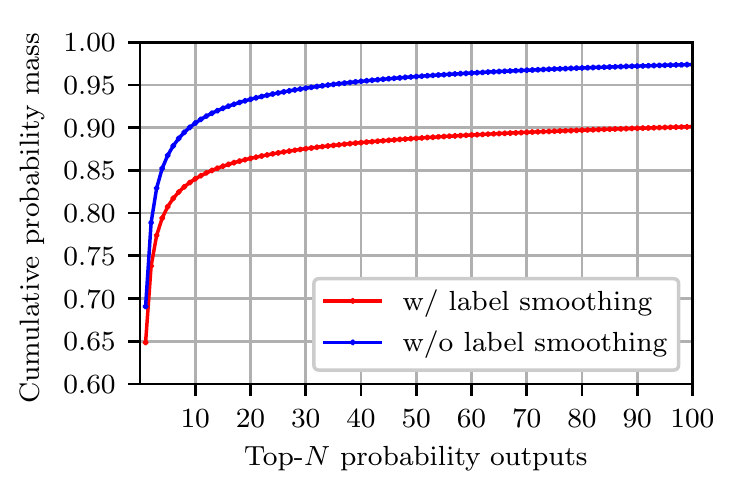}
}
\end{center}
\caption{Cumulative probabilities of the top-$N$ word candidates as estimated on newstest2015 English $\rightarrow$ German with and without label smoothing.
See section \ref{s:exp:controlled} for descriptions of the models.}
\label{fig:ls}
\end{figure}

To showcase the extent of this effect, we provide the average cumulative probabilities of top-$N$ words for NMT models, see Section \ref{s:exp:controlled}, trained with and without label smoothing in Figure \ref{fig:ls}.
The distributions are created on the development corpus.
We observe that training a model with label smoothing causes a re-allocation of roughly 7\% probability mass to all except the top-100 words.
This re-allocation is not problematic during beam search, since this strategy only looks at the top-scoring candidates.
However, when considering sampling for data generation, there is a high likelihood that one will sample from the space of low probability words, creating non-parallel outputs, see Table \ref{tbl:random-sample}.

\subsection{Restricting the Search Space}
\label{s:synthetic-data:search-space}

Changing the search approach $q$ is less arduous than changing the model $p$ since it does not involve re-training the model.
Restricting the search space to high-probability sentences avoids the issues highlighted in Section \ref{s:synthetic-data:issues} and provides a middle-ground between unrestricted sampling and beam search.

\newcite{edunov2018understanding} consider top-k sampling to avoid the aforementioned problem, however, there is no guarantee that the candidates are confident predictions.
We propose two alternative methods: (i) restrict the sampling outputs to words with a minimum probability and (ii) weighted sampling from the $N$-best candidates.

\subsubsection{Restricted Sampling}
\label{s:synthetic-data:search-space:r-sampling}

The first approach follows sampling directly from the model $p(\cdot | e_1^I, f_1^{j-1})$ at each position $j$, but only taking words with at least $\tau \in [0, 0.5)$ probability into account.
Afterwards, another softmax activation\footnote{Alternatively an L1-normalization would be sufficient.} is performed only over these words by masking all the remaining ones with large negative values.
If no words have over $\tau$ probability, then the maximum probability word is chosen.
Note that a large $\tau$ gets closer to greedy search ($\tau \geq 0.5$) and a lower value gets near to unrestricted sampling.

\begin{align}
&q(f | e_1^I, f_1^{j-1}; p) = \\
    &\begin{cases}
                    \text{softmax} \big( p(f | e_1^I, f_1^{j-1}), C \big), & {|C| > 0}  \vspace{5pt}\\ 
                    1, & \hspace{-5em} { \parbox[t]{.45\textwidth}{$|C| = 0 \wedge \\ f = \underset{f'}{\text{argmax}} \Big \{ p(f'| e_1^{I}, f_1^{j-1}) \Big \}$}}  \\
                    0, & \hspace{-5em} \text{ otherwise} \\
                    \end{cases} \nonumber
\end{align}
with $C \subseteq V_f$ being the subset of words of the source vocabulary $V_f$ with at least $\tau$ probability: 
\begin{align}
C= \big \{f ~ | ~p(f | e_1^I, f_1^{j-1}) \geq \tau \big \}
\end{align}
and $\text{softmax} \big( p(f | e_1^I, f_1^{j-1}), C \big)$ being a soft-max normalization restricted to the elements in $C$.

\subsubsection{$N$-best List Sampling}
\label{s:synthetic-data:search-space:n-best-sampling}

The second approach involves generating a list of $N$-best candidates, normalizing the output scores with a soft-max operation, as in Section \ref{s:synthetic-data:search-space:r-sampling}, and finally sampling a hypothesis. 

The score of a translation is abbreviated by $s(f_1^J | e_1^I) = \frac{1} {J} \log p(f_1^J | e_1^I)$.
\begin{align}
q_\text{nbest} &(f_1^J | e_1^I; p) = \\
&\begin{cases}
                     \text{softmax} \big( s(f_1^J | e_1^I), C \big), & f_1^J \in C \\
                    0, & \text{otherwise } \\
                    \end{cases} \nonumber
\end{align}
with $C \subseteq \mathbb{D}_{src}$ being the set of $N$-best translations found by the target-to-source model and $\mathbb{D}_{src}$ being the set of all source sentences:
\begin{align}
C = \underset{\mathcal{D} \subset \mathbb{D}_{src}: |\mathcal{D}| = N}{\text{argmax}} \Big \{ \sum_{f_1^J \in \mathcal{D}} s(f_1^J | e_1^I) \Big \} .
\end{align}

\section{Experiments}
\label{s:exp}

\subsection{Setup}
\label{s:exp:setup}

This section makes use of the WMT 2018 German $\leftrightarrow$ English \footnote{\url{http://www.statmt.org/wmt18/translation-task.html}} news translation task, consisting of 5.9M bilingual sentences. 
The German and English monolingual data is subsampled from the deduplicated NewsCrawl2017 corpus.
In total 4M sentences are used for German and English monolingual data.
All data is tokenized, true-cased and then preprocessed with joint byte pair encoding \cite{sennrich2016neural}\footnote{50k merge operations and a vocabulary threshold of 50 are used.}.

We train Base Transformer \cite{vaswani2017attention} models using the Sockeye toolkit \cite{hieber2017sockeye}.
Optimization is done with Adam \cite{kingma2014adam} with a learning rate of 3e-4, multiplied with 0.7 after every third 20k-update checkpoint without improvements in development set perplexity.
In Sections \ref{s:exp:controlled} and \ref{s:exp:real-world}, word batch sizes of 16k and 4k are used respectively.
Inference uses a beam size of $5$ and applies hypothesis length normalization.

Case-sensitive \bleu~\cite{papineni2002bleu} is computed using the \texttt{mteval\_13a.pl} script from Moses \cite{koehn2007moses}.
Model selection is performed based on the \bleu~performance on newstest2015.
All experiments were performed using the workflow manager Sisyphus \cite{peter2018sisyphus}.
We report the statistical significance of our results with MultEval \cite{clark2011better}.
A low p-value indicates that the performance gap between two systems is likely to hold given a different sample of a random process, e.g. an initialization seed.

\begin{table}[t]
\centering
\addtolength{\tabcolsep}{-3pt}
\begin{tabular}{l ccc}
\toprule
                  & test2015 & test2017 & test2018   \\ \midrule
beam search       & \ssig{30.9} & \ssig{31.9} & \textbf{40.1}  \\ \midrule
sampling          & \ssig{30.4} & \ssig{31.0} & \ssig{37.9}  \\ 
~~ w/o LS         & \ssig{30.4} & \ssig{31.3} & \ssig{37.9}  \\ 
~~ $\tau = 10\%$  & \ssig{\textbf{31.1}} & \ssig{\textbf{32.1}}  & 39.8  \\ 
50-best sampling  & \ssig{\textbf{31.1}} & \ssig{31.9}  & 39.8  \\ \midrule
reference         & \phantom{,}\textbf{32.6} & \phantom{,}\textbf{33.5} & 40.0  \\ \bottomrule
\end{tabular}
\caption{\bleupercent~results for the controlled scenario. \\
\ssig{} denotes a p-value of $<0.01$ w.r.t. the reference.}
\label{tbl:controlled-translation-results}

\end{table}

\subsection{Controlled Scenario}
\label{s:exp:controlled}

To compare the performance of each generation method to natural sentences, we shuffle and split the German $\rightarrow$ English bilingual data into 1M bilingual sentences and 4.9M monolingual sentences.
This gives us a reference translation for each sentence and eliminates domain adaptation effects.
The generator model is trained on the smaller corpus until convergence on \bleu, roughly 100k updates.
The final source-to-target model is trained from scratch on the concatenated synthetic and natural corpora until convergence on \bleu, roughly 250k updates for all variants.

Table \ref{tbl:controlled-translation-results} showcases the translation quality of the models trained on different kinds of synthetic corpora. 
Contrary to the observations in \newcite{edunov2018understanding}, unrestricted sampling does not outperform beam search and once the search space is restricted all methods perform similarly well.

To further investigate this, we look at other relevant statistics of a generated corpus and the performance of the subsequent models in Table \ref{tbl:controlled-metrics}.
These are the perplexities (\ppl) of the model on the training and development data and the entropy of a target-to-source IBM-1 model \cite{brown1993mathematics} trained with GIZA++ \cite{och2003systematic}.
The training set \ppl~varies strongly with each generation method since each produces hypotheses of differing quality.
All methods with a restricted search space have a larger translation entropy and smaller training \ppl~than natural data.
This is due to the sentences being less noisy and also the translation options being less varied.
Unrestricted sampling seems to overshoot the statistics of natural data, attaining higher entropy values.

However, once LS is removed, the best \ppl~on the development set is reached and the remaining statistics match the natural data very closely.
Nevertheless, the performance in \bleu~lags behind the methods that consider high-quality hypotheses as reported in Table \ref{tbl:controlled-translation-results}.
Looking further into the models, we notice that when trained on corpora with more variability, i.e. larger translation entropy, the probability distributions are flatter. 
We explain the better dev perplexities with unrestricted sampling with the same reason for which label smoothing is helpful: it makes the model less biased towards more common events \cite{ott2018analyzing}.
This uncertainty is, however, not beneficial for translation performance.

\begin{table}[t]
\centering
\addtolength{\tabcolsep}{-3pt}
\begin{tabular}{l ccc}
\toprule
     & Entropy & \multicolumn{2}{c}{\ppl} \\ 
     & {En $\rightarrow$ De} & Train & test2015 \\ \midrule
beam search         & 2.60 & 2.74 & 5.77 \\ \midrule
sampling            & 3.13 & 9.07 & 5.55 \\
~~ w/o LS           & \textbf{2.93} & \textbf{5.17} & \textbf{5.31}  \\ 
~~ $\tau = 10\%$    & 2.66 & 3.34 & 5.61 \\ 
50-best sampling    & 2.62 & 2.84 & 5.70 \\ \midrule
reference           & \textbf{2.91} & \textbf{5.18} & \textbf{4.50} \\ \bottomrule
\end{tabular}
\caption{IBM-1 model entropy and perplexity (\ppl) on the training and development set for the controlled scenario using different synthetic generation methods.}
\label{tbl:controlled-metrics}

\end{table}

\subsection{Real-world Scenario}
\label{s:exp:real-world}

Previously, we applied different synthetic data generation methods to a controlled scenario for the purpose of investigation. 
We extend the experiments to the original WMT 2018 German $\leftrightarrow$ English task and showcase the results in Table \ref{t:final:de-en}.
In contrast to the experiments of Section \ref{s:exp:controlled}, the distribution of the monolingual data now differs from the bilingual data.
The models are trained on the bilingual data for 1M updates and then fine-tuned for further 1M updates on the concatenated bilingual and synthetic corpora. 

The restricted sampling techniques perform comparable to or better than the other synthetic data generation methods in all cases.
Especially on English $\rightarrow$ German, unrestricted sampling only produces statistical significant improvements over beam search when LS is replaced.
Furthermore, restricting the search space via $50$-best list sampling improves significantly in both test sets.

We observe that on German $\rightarrow$ English newstest2018 particularly, there is a large drop in performance when using unrestricted sampling.
This is slightly alleviated by applying a minimum probability threshold of $\tau = 10\%$, but there is still a gap to be closed.
This behaviour is investigated in the following section.

\begin{figure*}[t]
\begin{center}
\scalebox{1.0}{
\includegraphics{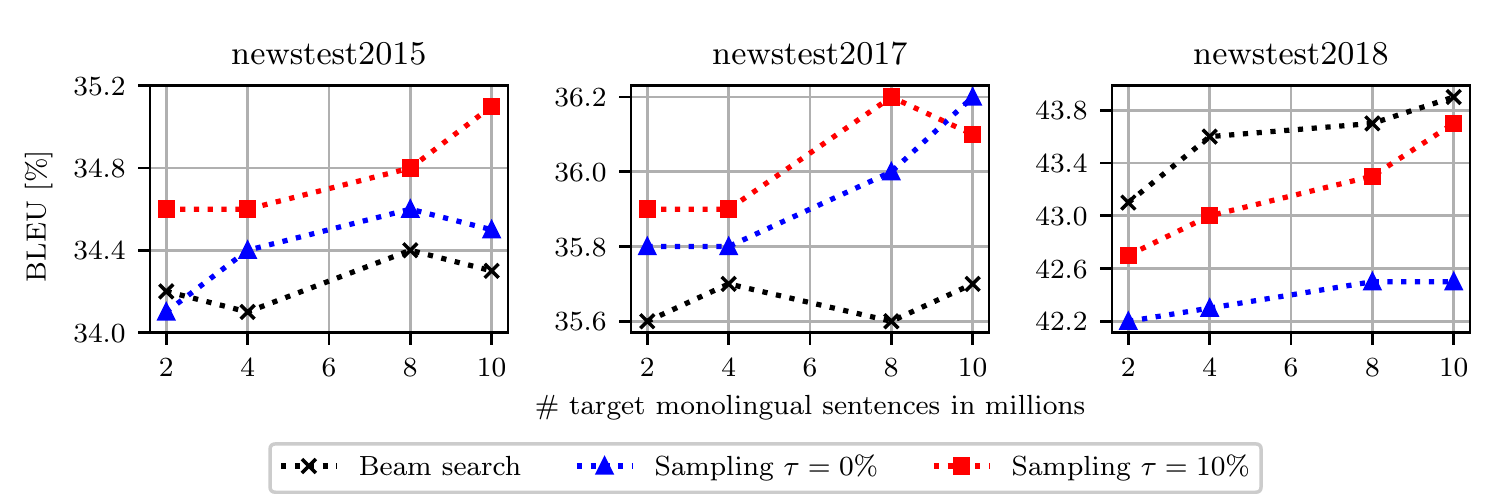}
}
\end{center}
\caption{WMT 2018 German $\rightarrow$ English \bleupercent~values comparing different synthetic data generation methods with a differing size of synthetic corpus.}
\label{fig:scalability}
\end{figure*}

\subsubsection{Scalability}
\label{s:exp:real-world:scalability}

A benefit of non-deterministic generation methods is the scalability in contrast to beam search. 
Under the assumption of a good fitting translation model, as argued in Section \ref{s:backtranslation}, sampling does appear to be the best option.

We compare different monolingual corpus sizes for the German $\rightarrow$ English task in Figure \ref{fig:scalability} on three different test sets.
Particularly, newstest2018 shows the exact opposite behaviour from the remaining test sets: the amount of data generated via beam search improves the resulting model, whereas sampling improves the system by a small margin.
Normal sampling has a general tendency to perform better with more data, but it saturates in two test sets (newstest2015 and newstest2018).
Restricted sampling appears to be the most consistent approach, always outperforming unrestricted sampling and also always scaling with a larger set of monolingual data.

These observations are strongly linked to the properties of current state-of-the-art models, see Section \ref{s:synthetic-data:issues} and experimental setup via e.g. the domain of the bilingual, monolingual and test data. 
Therefore, the high performance scaling with beam search in newstest2018 might be due to its \emph{relatedness} to the training data as measured by the high \bleu~values attained in inference.  

\begin{table}[t]
\centering
\addtolength{\tabcolsep}{-5pt}
\begin{tabular}{l ccccc} 
\toprule 
& \multicolumn{2}{c}{De $\rightarrow$ En} & \phantom{a} &  \multicolumn{2}{c}{En $\rightarrow$ De} \\ \cmidrule{2-3} \cmidrule{5-6} 
          & {\small test2017} & {\small test2018} && {\small test2017} & {\small test2018} \\ \midrule
beam search           & 35.7 & \textbf{43.6} && 28.2 & 41.3 \\ \midrule
sampling              & 35.8 & \ssig{42.3} && 28.6 & 41.5 \\ 
~~ w/o LS             & 35.9 & \ssig{42.5} && \ssig{\textbf{29.1}} & 41.7 \\ 
~~ $\tau = 10\%$      & 35.9 & \ssig{43.0} && \ssig{28.7} & 41.6 \\
$50$-best samp.       & \textbf{36.0} & \textbf{43.6} && \ssig{28.6} & \ssig{\textbf{41.8}} \\ \bottomrule 
\end{tabular}
\caption{WMT 2018 German $\leftrightarrow$ English \bleupercent~values comparing different synthetic data generation methods. 
\\\ssig{} denotes a p-value of $<0.01$ w.r.t. beam search.}
\label{t:final:de-en}

\end{table}

\subsection{Synthetic Source Examples}
\label{a:examples}

To highlight the issues present in unrestricted sampling, we compare the outputs of different generation methods in Table \ref{tbl:random-sample}. 
The unrestricted sampling output hypothesizes a second sentence which is not related at all to the input sentence but generates a much longer sequence. 
The restricted sampling methods and the model trained without label smoothing provide an accurate translation of the input sentence. 
Compared to the beam search hypothesis, they have a reasonable variation which is indeed closer to the human-translated reference.

\begin{table*}[t]
\centering
\begin{tabular}{l p{11.5cm}}
  \toprule
  source & it is seen as a long sag@@ a full of surprises . \\ \midrule 
beam search         &es wird als eine lange Geschichte voller \"{U}berraschungen angesehen . \\ \midrule
sampling            & es wird als eine lange S@@ aga voller \"{U}berraschungen angesehen . injury , Skepsis , Feuer ) , Duschen verursach@@ ter K\"{o}rper , Pal@@ \"{a}@@ ste , Gol@@ fen , Flu@@ r und Mu@@ ffen , Diesel@@ - Total Bab@@ ylon , der durch@@ s Wasser und Wasser@@ kraft fließt .\\ \midrule
~~ w/o label smoothing & es wurde als eine lange Geschichte voller \"{U}berraschungen gesehen . \\  \midrule
~~ $\tau = 10\%$    & es wird als lange S@@ age voller \"{U}berraschungen angesehen . \\  \midrule
50-best sampling    & es wird als eine lange S@@ age voller \"{U}berraschungen gesehen . \\ \midrule
reference  & er wird als eine lange S@@ aga voller \"{U}berraschungen angesehen . \\ \bottomrule
\end{tabular}
\caption{Random example generated by different methods for the controlled scenario of WMT 2018 German $\rightarrow$ English. @@ denotes the subword token delimiter.}
\label{tbl:random-sample}
\end{table*}

\section{Conclusion}
\label{s:conclusion}

In this work, we link the optimization criterion of an NMT model with a synthetic data generator defined for both beam search and additional sampling-based methods.
By doing so, we identify that the search method plays an important role, as it is responsible for offsetting the shortcomings of the generator model.
Specifically, label smoothing and probability smearing issues cause sampling-based methods to generate unnatural sentences.

We analyze the performance of our techniques on a closed- and open-domain of the WMT 2018 German $\leftrightarrow$ English news translation task.
We provide qualitative and quantitative evidence of the detrimental behaviours and show that these can be influenced by re-training the generator model without label smoothing or by restricting the search space to not consider low-probability outputs.
In terms of translation quality, sampling from $50$-best lists outperforms beam search, albeit at a higher computational cost.
Restricted sampling or the disabling of label smoothing for the generator model are shown to be cost-effective ways of improving upon the unrestricted sampling approach of \newcite{edunov2018understanding}.

\section*{Acknowledgments}

\begin{center}
\vspace{0.5em}
\includegraphics[width=0.2\textwidth]{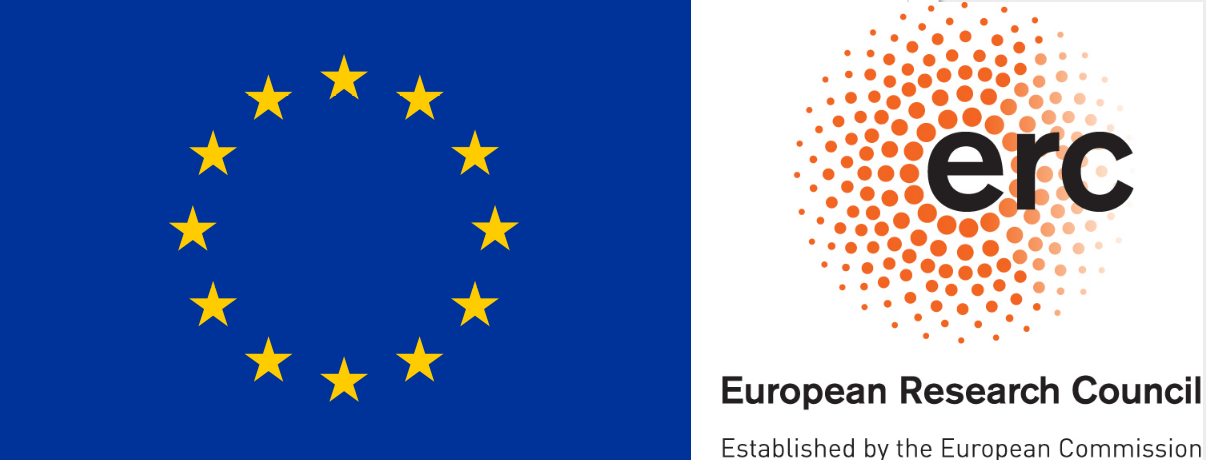}
\hspace{6pt}
\includegraphics[width=0.09\textwidth]{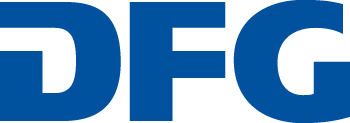}
\hspace{4pt}
\includegraphics[width=0.13\textwidth]{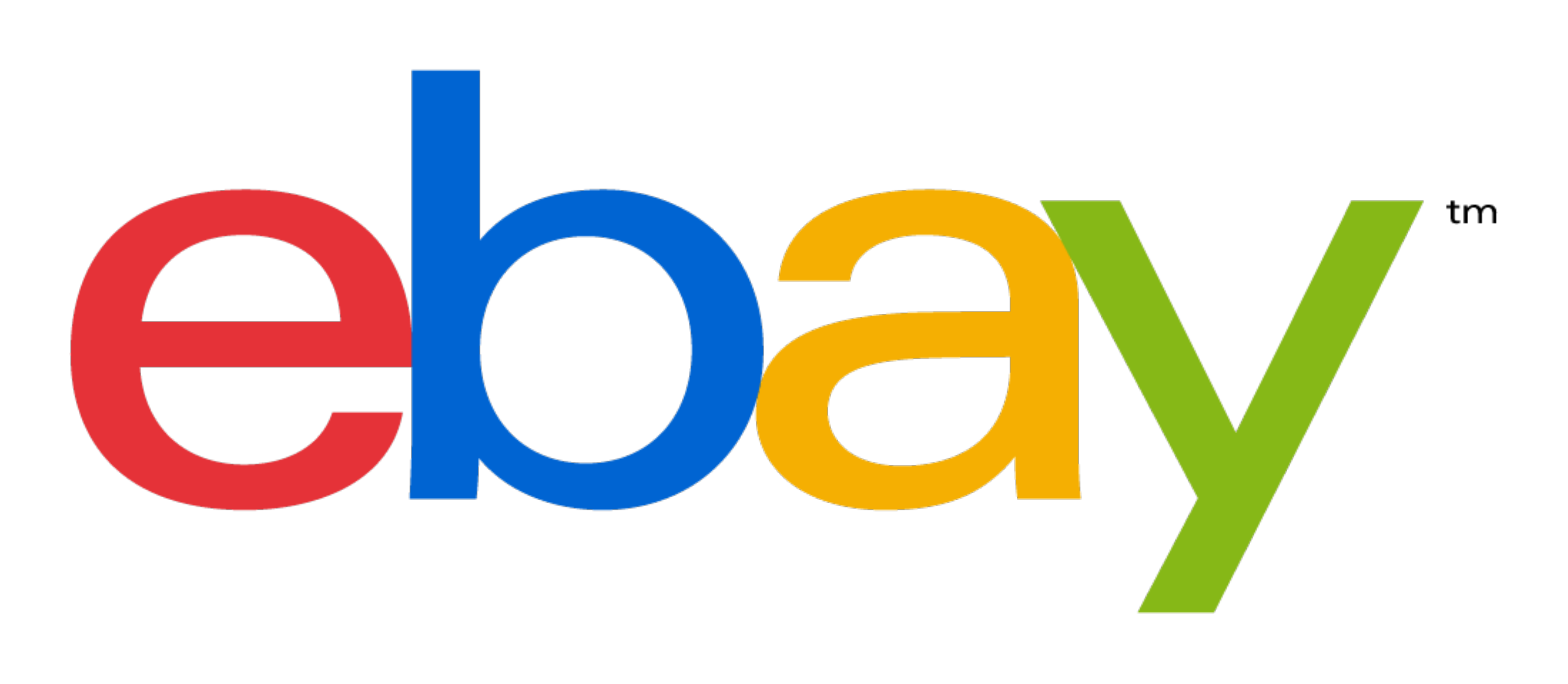}
\end{center}
\vspace{0.5em}

This work has received funding from the European Research Council (ERC) (under the European Union's Horizon 2020 research and innovation programme, grant agreement No 694537, project "SEQCLAS"), the Deutsche Forschungsgemeinschaft (DFG; grant agreement NE 572/8-1, project "CoreTec"), and eBay Inc. The GPU cluster used for the experiments was partially funded by DFG Grant INST 222/1168-1. The work reflects only the authors' views and none of the funding agencies is responsible for any use that may be made of the information it contains.

\bibliographystyle{acl_natbib}
\bibliography{acl2019}

\begin{thebibliography}{20}
\expandafter\ifx\csname natexlab\endcsname\relax\def\natexlab#1{#1}\fi

\bibitem[{Bahdanau et~al.(2014)Bahdanau, Cho, and Bengio}]{bahdanau2014neural}
Dzmitry Bahdanau, Kyunghyun Cho, and Yoshua Bengio. 2014.
\newblock Neural machine translation by jointly learning to align and
  translate.
\newblock \emph{arXiv preprint arXiv:1409.0473}.
\newblock Version 4.

\bibitem[{Brown et~al.(1993)Brown, Pietra, Pietra, and
  Mercer}]{brown1993mathematics}
Peter~F Brown, Vincent J~Della Pietra, Stephen A~Della Pietra, and Robert~L
  Mercer. 1993.
\newblock The mathematics of statistical machine translation: Parameter
  estimation.
\newblock \emph{Computational linguistics}, 19(2):263--311.

\bibitem[{Burlot and Yvon(2018)}]{burlot2018using}
Franck Burlot and Fran{\c{c}}ois Yvon. 2018.
\newblock Using monolingual data in neural machine translation: a systematic
  study.
\newblock In \emph{Proceedings of the Third Conference on Machine Translation
  (WMT 2018)}, pages 144--155.

\bibitem[{Clark et~al.(2011)Clark, Dyer, Lavie, and Smith}]{clark2011better}
Jonathan~H Clark, Chris Dyer, Alon Lavie, and Noah~A Smith. 2011.
\newblock Better hypothesis testing for statistical machine translation:
  Controlling for optimizer instability.
\newblock In \emph{Proceedings of the 49th Annual Meeting of the Association
  for Computational Linguistics (ACL 2011)}, pages 176--181.

\bibitem[{Cotterell and Kreutzer(2018)}]{cotterell2018explaining}
Ryan Cotterell and Julia Kreutzer. 2018.
\newblock Explaining and generalizing back-translation through wake-sleep.
\newblock \emph{arXiv preprint arXiv:1806.04402}.
\newblock Version 1.

\bibitem[{Edunov et~al.(2018)Edunov, Ott, Auli, and
  Grangier}]{edunov2018understanding}
Sergey Edunov, Myle Ott, Michael Auli, and David Grangier. 2018.
\newblock Understanding back-translation at scale.
\newblock \emph{arXiv preprint arXiv:1808.09381}.
\newblock Version 2.

\bibitem[{Hieber et~al.(2017)Hieber, Domhan, Denkowski, Vilar, Sokolov,
  Clifton, and Post}]{hieber2017sockeye}
Felix Hieber, Tobias Domhan, Michael Denkowski, David Vilar, Artem Sokolov, Ann
  Clifton, and Matt Post. 2017.
\newblock Sockeye: A toolkit for neural machine translation.
\newblock \emph{arXiv preprint arXiv:1712.05690}.
\newblock Version 2.

\bibitem[{Hinton et~al.(1995)Hinton, Dayan, Frey, and Neal}]{hinton1995wake}
Geoffrey~E Hinton, Peter Dayan, Brendan~J Frey, and Radford~M Neal. 1995.
\newblock The" wake-sleep" algorithm for unsupervised neural networks.
\newblock \emph{Science}, 268(5214):1158--1161.

\bibitem[{Hoang et~al.(2018)Hoang, Koehn, Haffari, and
  Cohn}]{hoang2018iterative}
Vu~Cong~Duy Hoang, Philipp Koehn, Gholamreza Haffari, and Trevor Cohn. 2018.
\newblock Iterative back-translation for neural machine translation.
\newblock In \emph{Proceedings of the 2nd Workshop on Neural Machine
  Translation and Generation (WNMT 2018)}, pages 18--24.

\bibitem[{Imamura et~al.(2018)Imamura, Fujita, and
  Sumita}]{imamura2018enhancement}
Kenji Imamura, Atsushi Fujita, and Eiichiro Sumita. 2018.
\newblock Enhancement of encoder and attention using target monolingual corpora
  in neural machine translation.
\newblock In \emph{Proceedings of the 2nd Workshop on Neural Machine
  Translation and Generation (WNMT 2018)}, pages 55--63.

\bibitem[{Kingma and Ba(2014)}]{kingma2014adam}
Diederik~P Kingma and Jimmy Ba. 2014.
\newblock Adam: A method for stochastic optimization.
\newblock \emph{arXiv preprint arXiv:1412.6980}.
\newblock Version 9.

\bibitem[{Koehn et~al.(2007)Koehn, Hoang, Birch, Callison-Burch, Federico,
  Bertoldi, Cowan, Shen, Moran, Zens et~al.}]{koehn2007moses}
Philipp Koehn, Hieu Hoang, Alexandra Birch, Chris Callison-Burch, Marcello
  Federico, Nicola Bertoldi, Brooke Cowan, Wade Shen, Christine Moran, Richard
  Zens, et~al. 2007.
\newblock Moses: Open source toolkit for statistical machine translation.
\newblock In \emph{Proceedings of the 45th Annual Meeting of the Association
  for Computational Linguistics (ACL 2007)}, pages 177--180.

\bibitem[{Och and Ney(2003)}]{och2003systematic}
Franz~Josef Och and Hermann Ney. 2003.
\newblock A systematic comparison of various statistical alignment models.
\newblock \emph{Computational linguistics}, pages 19--51.

\bibitem[{Ott et~al.(2018)Ott, Auli, Granger, and Ranzato}]{ott2018analyzing}
Myle Ott, Michael Auli, David Granger, and Marc'Aurelio Ranzato. 2018.
\newblock Analyzing uncertainty in neural machine translation.
\newblock \emph{arXiv preprint arXiv:1803.00047}.
\newblock Version 4.

\bibitem[{Papineni et~al.(2002)Papineni, Roukos, Ward, and
  Zhu}]{papineni2002bleu}
Kishore Papineni, Salim Roukos, Todd Ward, and Wei-Jing Zhu. 2002.
\newblock Bleu: a method for automatic evaluation of machine translation.
\newblock In \emph{Proceedings of the 40th Annual Meeting on Association for
  Computational Linguistics (ACL 2002)}, pages 311--318.

\bibitem[{Pereyra et~al.(2017)Pereyra, Tucker, Chorowski, Kaiser, and
  Hinton}]{pereyra2017regularizing}
Gabriel Pereyra, George Tucker, Jan Chorowski, {\L}ukasz Kaiser, and Geoffrey
  Hinton. 2017.
\newblock Regularizing neural networks by penalizing confident output
  distributions.
\newblock \emph{arXiv preprint arXiv:1701.06548}.
\newblock Version 1.

\bibitem[{Peter et~al.(2018)Peter, Beck, and Ney}]{peter2018sisyphus}
Jan-Thorsten Peter, Eugen Beck, and Hermann Ney. 2018.
\newblock Sisyphus, a workflow manager designed for machine translation and
  automatic speech recognition.
\newblock In \emph{Proceedings of the 2018 Conference on Empirical Methods in
  Natural Language Processing (EMNLP 2018)}, pages 84--89.

\bibitem[{Sennrich et~al.(2016{\natexlab{a}})Sennrich, Haddow, and
  Birch}]{sennrich2016improving}
Rico Sennrich, Barry Haddow, and Alexandra Birch. 2016{\natexlab{a}}.
\newblock Improving neural machine translation models with monolingual data.
\newblock In \emph{Proceedings of the 54th Annual Meeting of the Association
  for Computational Linguistics (ACL 2016)}, pages 86--96.

\bibitem[{Sennrich et~al.(2016{\natexlab{b}})Sennrich, Haddow, and
  Birch}]{sennrich2016neural}
Rico Sennrich, Barry Haddow, and Alexandra Birch. 2016{\natexlab{b}}.
\newblock Neural machine translation of rare words with subword units.
\newblock In \emph{Proceedings of the 54th Annual Meeting of the Association
  for Computational Linguistics (ACL 2016)}, pages 1715--1725.

\bibitem[{Vaswani et~al.(2017)Vaswani, Shazeer, Parmar, Uszkoreit, Jones,
  Gomez, Kaiser, and Polosukhin}]{vaswani2017attention}
Ashish Vaswani, Noam Shazeer, Niki Parmar, Jakob Uszkoreit, Llion Jones,
  Aidan~N Gomez, {\L}ukasz Kaiser, and Illia Polosukhin. 2017.
\newblock Attention is all you need.
\newblock In \emph{Advances in Neural Information Processing Systems (NIPS
  2017)}, pages 6000--6010.

\end{thebibliography}

\end{document}